\documentclass[10pt,twocolumn,letterpaper]{article}

\usepackage{times}
\usepackage{epsfig}
\usepackage{graphicx}
\usepackage{amsmath}
\usepackage{amssymb}
\usepackage{url}            
\usepackage{booktabs}       
\usepackage{amsfonts}       
\usepackage{nicefrac}       
\usepackage{microtype}      
\usepackage{commath}
\usepackage{wrapfig}
\usepackage{mathtools,amsbsy}

\usepackage[margin=20mm,columnsep=7mm]{geometry}

\usepackage[inline]{enumitem} 

\usepackage[all]{foreign}   

\usepackage[pagebackref=true,breaklinks=true,colorlinks,bookmarks=false]{hyperref}

\makeatletter
\def\maketag@@@#1{\hbox{\m@th\normalfont\normalsize#1}}
\makeatother

\begin{document}

\title{\textbf{Embed Me If You Can: A Geometric Perceptron}}

\author{%
	Pavlo Melnyk,\quad Michael Felsberg,\quad Mårten Wadenbäck\\
	{\small Computer Vision Laboratory, Department of Electrical Engineering, Linköping University}\\
	{\small \texttt{\{pavlo.melnyk, michael.felsberg, marten.wadenback\}@liu.se}}
}

\date{}

\maketitle

\begin{abstract}
Solving geometric tasks involving point clouds by using machine learning is a challenging problem. Standard feed-forward neural networks combine linear or, if the bias parameter is included, affine layers and activation functions. Their geometric modeling is limited, which motivated the prior work introducing the multilayer hypersphere perceptron (MLHP). 
Its constituent part, \ie, the hypersphere neuron, is obtained by applying a conformal embedding of Euclidean space. By virtue of Clifford algebra, it can be implemented as the Cartesian dot product of inputs and weights. 
If the embedding is applied in a manner consistent with the dimensionality of the input space geometry, the decision surfaces of the model units become combinations of hyperspheres and make the decision-making process geometrically interpretable for humans. 
Our extension of the MLHP model, the multilayer geometric perceptron (MLGP), and its respective layer units, \ie, geometric neurons, are consistent with the 3D geometry and provide a geometric handle of the learned coefficients. In particular, the geometric neuron activations are isometric in 3D, which is necessary for rotation and translation equivariance.
When classifying the 3D Tetris shapes, we quantitatively show that our model requires no activation function in the hidden layers other than the embedding to outperform the vanilla multilayer perceptron. 
In the presence of noise in the data, our model is also superior to the MLHP.
\end{abstract}

\section{Introduction}

Understanding the geometry of a neuron is a crucial prerequisite to successfully performing geometric deep learning~\cite{bronstein2017geometric}.
Owing to their inherent connection to the notion of \emph{distance}, circles (or spheres) are fundamental atomic structures for defining geometric constraints.
Note, \eg, how the third corner of a triangle is constrained by specifying the radii of two circles centered in each of the other two corners.

This motivates us to consider geometries of decision surfaces beyond hyperplanes, such as hyperspheres, in order to properly represent isometries.

Going beyond planar decision surfaces to non-planar ones increases the complexity of the model, but this has proved to be beneficial performance-wise \cite{buchholz2000hyperbolic, banarer2003hypersphere, lipson2000clustering}.
To construct and represent such surfaces, one needs to consider more general spaces, in which case, Klein geometries \cite{sharpe2000differential} provide a useful theoretical framework.

In this paper, we employ Clifford algebra as a tool to perform computations in conformal geometry. Building on top of the previous works on Clifford neurons \cite{buchholz2008clifford} and spherical decision surfaces \cite{perwass2003spherical}, we explore the multilayer hypersphere perceptron (MLHP) \cite{banarer2003design} model applied to the problem of classifying the 3D Tetris shapes, essentially a collection of point clouds, see Section~\ref{shape_classification} and Fig.~\ref{fig:tetris}. To the best of our knowledge, MLHP has not been previously investigated in this context.
We focus specifically on 3D geometry, which is important for tasks such as pose estimation, which in turn is a prerequisite for grasping, 3D inpainting, and augmented reality.

Striving to make the decision-making process more intuitive and be consistent with the dimensionality of the input space geometry, we perform the conformal embedding in a Minkowski space.  
Consequently, we observe that the decision surfaces of the MLHP units become combinations of hyperspheres. We call such an extension of the MLHP model the \emph{multilayer geometric perceptron} (MLGP) and refer to its respective layer units as \emph{geometric neurons}.
Owing to the homogeneous representation \cite{hartley2003multiple}, we provide an interpretation of our model parameters \emph{directly} in the Euclidean space, which is only intuitive if the embedding agrees with the input geometry.

\noindent We summarize our contributions as follows:
\begin{figure*}
	\centering
	\includegraphics[width=0.7\linewidth]{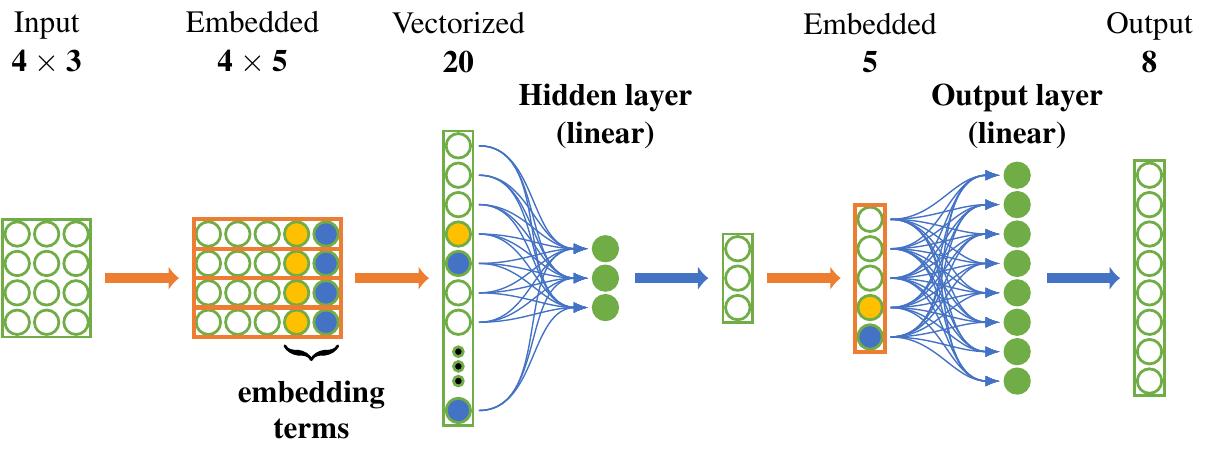}
	\caption{The proposed (feed-forward) MLGP model --- our modification of the baseline MLHP. The embedding of the input array is performed \textit{point-wise} and at each subsequent layer vector-wise: the first embedding term (yellow) is always set to $-1$, the second (blue) is the scaled magnitude of the vector being embedded, \ie, $-\frac{1}{2}||\textbf{x}||^2$. Since the embedding of the model input is point-wise, the hidden layer consists of \textit{geometric neurons} that represent combinations of hyperspheres. The output layer consists of hypersphere neurons \cite{banarer2003hypersphere}.}
	\label{fig:model}
\end{figure*}

\vspace{-2.5mm}

\begin{enumerate}[label=(\textbf{\alph*}),itemsep=-3.5pt] 
 	\item We demonstrate how spherical decision surfaces improve the understanding of the decision-making process of neural networks, provided that their construction is done adhering to the input space geometry.

	\item We propose an extension to the MLHP, the MLGP model\footnote{The code is available at \href{https://github.com/pavlo-melnyk/mlgp-embedme}{github.com/pavlo-melnyk/mlgp-embedme}.} with \emph{ geometric neurons}, and show that, apart from being more geometrically explainable, it achieves favorable quantitative results when classifying  the 3D Tetris shapes, and is superior when they are perturbed.

	\item By introducing a matrix operator, we derive the isomorphism between the sandwich product of the motor (rotation followed by translation) with a general geometric object in the conformal $\mathbb{ME}^3\equiv\mathbb{R}^{3+1, 1}$ space (see Section~\ref{sec:minkowski embedding} for notation) and the corresponding matrix--vector product in the Euclidean $\mathbb{R}^5$ space. 
	Using this result, we prove that the geometric neuron activations are isometric in $\mathbb{R}^3$ by construction, which is necessary for rotation and translation equivariance. We further demonstrate it experimentally.

\end{enumerate}

\section{Related work}
\label{related work}
Research on neural networks equivariant to certain symmetry groups has been expanding over the past few years, \eg, SE(3)-equivariant models \cite{weiler20183d}, \cite{thomas2018tensor}, and the SO(3)-equivariant network \cite{anderson2019cormorant}. 
To introduce to the reader a broader picture of concepts related to our work, we start by noting that a Klein geometry can be viewed as a homogeneous space together with a transformation (symmetry) group acting on that space. Conformal geometry on the sphere is modeled as a Klein geometry with the underlying space being the sphere $\mathcal{S}^{n}$  and the \textit{Lorentz group} \cite{li2001generalized} of an $(n+2)$-dimensional space (\eg, $\mathbb{R}^{n+1,1}$) acting as the transformation group.  The main computational mechanism in conformal geometry --- Clifford (geometric) algebras --- is utilized in all the following methods.

\subsection{Hyperspherical decision surfaces}
We draw the main inspiration for our work from the idea of modeling hyperspherical surfaces using a conformal space representation introduced in \cite{li2001universal} and exploited in \cite{perwass2003spherical}. The hypersphere neuron with, as the name suggests, a hypersphere as decision surface is proposed as a variant of a Clifford neuron in \cite{banarer2003hypersphere}. Therein, a hyperspherical surface is shown as a generalization of a hyperplane.
A multilayer feed-forward neural network based on hypersphere neurons is designed in \cite{banarer2003design} and is referred to as MLHP. The authors describe how a certain amount of reduction in computational complexity can be achieved when using the MLHP  model for some types of learning tasks. However, the prior work did not consider the problem of classifying point clouds.

\subsection{Clifford neural networks}
A multilayer Clifford neural network, as well as the corresponding back-propagation derivation, is first proposed and discussed in \cite{Pearson92backpropagation, pearson1994neural}.
As per \cite{buchholz2008clifford}, work on another multilayer model, two key concepts for Clifford neural machinery originate and are easy to analyze at the neuron level:
\begin{enumerate*}[label=(\textit{\roman*})]
	\item the ability to process various geometric entities, and
	\item the concept of the geometric model.
\end{enumerate*}
The latter acts as certain transformations on the processed data and becomes inherent by choosing a particular Clifford algebra.
The paper \cite{buchholz2008clifford} also introduces the spinor Clifford neurons (SCN) with weights acting like rotors from two sides. It is demonstrated how a single SCN
can be used to compute Möbius transformations in a linear way: something unattainable by any real-valued network.
Additionally, the paper describes Clifford-valued activation functions for all two-dimensional Clifford algebras.

The work \cite{serrano2017hyperconic} introduces the hyperconic multilayer perceptron with quadratic hypersurfaces spawned by the hyperconic neurons in the hidden layer. The output units in their model are hypersphere neurons. The model is trained using the particle swarm optimization (PSO) algorithm \cite{eberhart1995new}.
The authors of \cite{villasenor2017hyperellipsoidal} use the generalization of the geometric algebra of quadratic surfaces $\mathcal{G}_{6,3}$ \cite{zamora2014g} and propose a new Clifford neuron --- the hyperellipsoidal neuron.
It is shown how decision surfaces of different geometric shapes, derived as special cases of an ellipsoid, can be obtained, depending on the input data: spherical decision surface, ellipsoidal, cylindrical, or a pair of planes.
They use some hybrid training algorithm, in which the center of the hyperellipsoid is updated by unsupervised learning and the radii by a supervised learning method.

Less related to geometric problems are models of recurrent Clifford NNs originally discussed in \cite{kuroe2011models}, where their dynamics are studied from the perspective of the existence of energy functions.

We focus on spherical decision surfaces and feed-forward-like models trainable with backpropagation, but in contrast to prior work, we exploit a different embedding scheme that is consistent with the 3D geometry of the input space. Consequently, the first hidden layer units in our model can be seen as combinations of hypersphere neurons and, along with the rest of the layers, do not necessarily require an activation function. Moreover, such units activations are isometric to ridig body transformations of the input. Utilizing the chosen embedding strategy results in a more explainable decision-making process and can even be superior performance-wise when dealing with noisy data. We explain the details and discuss the advantages of our method in Section \ref{method}.

\section{Background}
\label{background}
The proposed spherical model MLGP, which we will present in detail in Section \ref{method}, is based on a particular embedding of the Euclidean vector space $\mathbb{R}^3$ into a Minkowski space, which is isomorphic to the Euclidean space $\mathbb{R}^5$.
The construction of this embedding is reminiscent of the way in which projective $n$-space is embedded in $\mathbb{R}^{n+1}$ by means of homogeneous coordinates.
Many key aspects pertaining to the geometric interpretation will also be similar for these two embeddings, and we therefore encourage the reader to keep the more familiar projective case in mind when studying our discussion of the conformal embedding in Section~\ref{sec:minkowski embedding}.

In particular, we wish to emphasize the fact that successful interpretation of various geometric entities, given in their embedding representation, requires appropriate normalization to be applied.
To make sense of the homogeneous coordinates of a (proper) point, for example, the appropriate \emph{point normalization} is achieved by simply dividing the whole coordinate vector by its final coordinate.
Correspondingly, recall that the dual projective entities (\ie,  hyperplanes) require a fundamentally different type of normalization; here we require that the final coordinate is $\leq 0$ and that the squares of the others sum to one.
If $\textbf{p}=(p_1,\ldots,p_n,-\Delta)$ is normalized in this way, it is possible to directly interpret $(p_1,\ldots,p_n)$ as the outward pointing unit normal of the hyperplane, and $\Delta \geq 0$ as its distance to the origin.

Slightly related, we also wish to highlight the role of scalar products between objects in the embedding space.
In the projective embedding, we recall that incidence relations between points and hyperplanes are expressed as orthogonality of the embedding vectors, \ie $\textbf{p}^\top\textbf{x}=0$.
If this scalar product is not zero, it does not have a direct geometric interpretation, \emph{unless the representations have been properly normalized}.
If proper normalization has been applied, then of course $\abs{\textbf{p}^\top\textbf{x}}$ is simply the Euclidean distance between the point and the hyperplane.

\subsection{Minkowski space and conformal embedding}\label{sec:minkowski embedding}
Taking one step further from homogeneous coordinates opens up a whole new world in the form of \textit{conformal geometry} \cite{li2001universal}, \ie, angle-preserving transformations on a space. 
\textit{Minkowski space} \cite{li2001generalized}, named after H. Minkowski who introduced $\mathbb{R}^{3, 1}$ as a model of space-time, is the real vector space $\mathbb{ME}^n\equiv\mathbb{R}^{n+1, 1}$ where the first $n+1$ basis vectors square to $+1$ and the last of them squares to $-1$. 

\textbf{Minkowski $\mathbb{R}^{1,1}$ space.\hspace{1em}}  The relevance of Minkowski spaces for Euclidean geometry is well-described in terms of the Minkowski $\mathbb{R}^{1, 1}$ plane in \cite{li2001generalized}.
Its \textit{orthonormal basis} is defined as $\{e_{+},~e_{-}\}$,  where $e_{+}^2 = 1$, ~$e_{-}^2 = -1$, and $e_{+} \cdot e_{-} = 0$. 

A \textit{ null basis} can then be constructed as the two vectors $\{e_{0},~e_{\infty}\}$, where
$e_{0} = \frac{1}{2} (e_{-} - e_{+})$ is the origin and $ e_{\infty} = e_{-} + e_{+}$ is the point at infinity.
Note the properties $e_{0}^2 = e_{\infty}^2 = 0$ and $e_{0} \cdot e_{\infty} = -1$, which follow from the signature properties of $e_{+}$ and $e_{-}$ and the fact that they are orthogonal.

\textbf{Conformal embedding.\hspace*{1em}} Given a vector in Euclidean space, $\textbf{x}\in\mathbb{R}^n$, one can construct the conformal space as $\mathbb{ME}^{n} \equiv \mathbb{R}^{n+1,1} = \mathbb{R}^{n} \oplus \mathbb{R}^{1,1}$.
The embedding of $\textbf{x}$ in the conformal space $\mathbb{ME}^n$ represents the stereographic projection of $\textbf{x}$ onto a projection sphere defined in $\mathbb{ME}^n$ as
\begin{equation}
\label{conformal_embedding}
X = \mathcal{C}(\textbf{x}) = \textbf{x} + \frac{1}{2}\textbf{x}^2~e_{\infty}+e_{0}~,
\end{equation}
where $X \in \mathbb{ME}^n$ is called \textit{normalized} and $\textbf{x}^2 = \textbf{x} \cdot \textbf{x} = ||\textbf{x}||^2$. Observe that $X^2=0$. Note in this context that the  \textit{Clifford (geometric) product} of vectors (see equation (1) in \cite{hitzer2013geometric}) is denoted as concatenation of literals. The standard inner product is related to the geometric product as
\begin{equation}
\label{eq:inner}
\textbf{x}\cdot\textbf{y}=\frac{\textbf{x}\textbf{y}+\textbf{y}\textbf{x}}{2}\quad\text{and}\quad X\cdot Y=\frac{XY+YX}{2}~.
\end{equation}

From \eqref{conformal_embedding}, we obtain the naming of the two null vectors:
\begin{equation}
	e_{0} = \mathcal{C}(\textbf{0})
	\quad\textup{and}\quad
	e_{\infty} = \lim\limits_{\abs{\textbf{x}} \rightarrow \infty} \frac{2}{\textbf{x}^2}~\mathcal{C}(\textbf{x}).
\end{equation}
The embedding \eqref{conformal_embedding} is homogeneous, \ie, all embedding vectors in the equivalence class
\begin{equation}
\label{homogeneous_embedding}
[X] = \big\{Z\in\mathbb{R}^{n+1, 1} : Z=\gamma X,\;\gamma\in\mathbb{R} \setminus\{0\}\big\}
\end{equation}
are taken to represent the same vector $\textbf{x}$.
This property is fundamental for the remainder of the paper.

\textbf{Scalar products.\hspace*{1em}} 
Given $Y=\textbf{y} + \frac{1}{2}\textbf{y}^2~e_{\infty}+e_{0}$, the scalar product of two embeddings in conformal space turns out to be the Euclidean distance, which constitutes the basis for deriving the hypersphere neuron \cite{banarer2003hypersphere}:
\begin{equation}
	X\cdot Y = -\frac{1}{2}(\textbf{x}-\textbf{y})^2~.
\end{equation}

\subsection{Hypersphere as classifier}
\label{hyperphere_as_classifier}
A \textit{normalized hypersphere} in $\mathbb{ME}^n$ is a hypersphere $S\in\mathbb{ME}^n$ with center $\textbf{c} =(c_1, \dots, c_n)\in\mathbb{R}^n$ embedded as $C\in\mathbb{ME}^n$, radius $r\in\mathbb{R}$,  and the coefficient for $e_{0}$ set to 1. It is defined in the conformal space as
\begin{equation}
	\label{eq:sphere_in_conformal_space}
	S = \textbf{c} + \frac{1}{2}(\textbf{c}^2-r^2)~e_{\infty} + e_{0} = C - \frac{1}{2}r^2 e_{\infty}~.
\end{equation}

Keeping in mind that $\mathbb{R}^n$ has a basis $(e_1, \dots, e_n)$, we obtain the scalar product of an embedded data vector $X$ and a hypersphere $S$ in $\mathbb{ME}^n$:
\begin{equation}
	\label{eq:scalar_product_in_conformal_space_euclidean_interpretation}
	X\cdot S = X\cdot C - \frac{1}{2}r^2 X \cdot e_{\infty} =  -\frac{1}{2}(\textbf{x}-\textbf{c})^2 + \frac{1}{2}r^2~.
\end{equation}

Thus, it follows that $X\cdot S = 0 \iff \abs{\textbf{x} - \textbf{c}} = \abs{r}$. Specifically, the scalar product shows where the input vector is relative to the hypersphere: inside (positive product), on (zero), or outside (negative product) of the hypersphere.

It has been shown in \cite{banarer2003hypersphere} that by embedding a data vector $\textbf{x} = (x_1, \dots, x_n)\in\mathbb{R}^n$ and the hypersphere $S\in\mathbb{ME}^n$ in $\mathbb{R}^{n+2}$ as
\begin{equation}
\label{hypersphere_in_r}
	\begin{aligned}
		\textbf{\textit{X}} &= (x_1, \dots, x_n, -1, -\frac{1}{2}\textbf{x}^2)\in\mathbb{R}^{n+2},\\
		\textbf{\textit{S}} &= (c_1, \dots, c_n, \frac{1}{2}(\textbf{c}^2 - r^2), 1)\in\mathbb{R}^{n+2},
	\end{aligned}
\end{equation}
with $\textbf{\textit{S}}$ further referred to as a \textit{normalized hypersphere} (in $\mathbb{R}^{n+2}$), one can implement a hypersphere neuron in $\mathbb{ME}^n$ as a standard dot product in $\mathbb{R}^{n+2}$ since
\begin{equation}
	\label{scalar_products_connection}
	\begin{aligned}
		\textbf{\textit{X}} \cdot \textbf{\textit{S}}
		&= \textbf{x}\cdot\textbf{c}~-\frac{1}{2}(\textbf{c}^2 - r^2)-\frac{1}{2}\textbf{x}^2 = \\
		&= -\frac{1}{2}(\textbf{x} - \textbf{c})^2 +\frac{1}{2}r^2
		= X\cdot S ~.
	\end{aligned} 
\end{equation}

\section{The proposed MLGP model}
\label{method}
From now on and depending on the context, \textit{hypersphere} refers to either a decision surface (geometric entity), or the (scaled) embedded vector $\textbf{\textit{S}}\in\mathbb{R}^{n+2}$ (\ref{hypersphere_in_r}), or a classifier (the hypersphere neuron).

\subsection{Geometric neuron: point-wise embedding}
\label{design_choice}

In the MLHP \cite{banarer2003design}, the model input 
is treated as 
a single real $n$-vector 
that is subsequently embedded in $\mathbb{R}^{n+2}$, as discussed in Section~\ref{hyperphere_as_classifier}. However, such an embedding scheme is not the most intuitive choice for all types of learning problems.

We propose to apply the conformal embedding to the input \textit{point-wise}, in contrast to performing the embedding on a vectorized input as in MLHP \cite{banarer2003design}. We motivate this choice by reasoning from a geometric perspective.

For the simplicity of argument, consider 
3D geometry. Take, e.g., the 
geometry problem of classifying point clouds, each consisting of $k$ points, to which random 3D rigid transformations are applied. 
Suppose one applies the transformation to 
the points $\mathbb{R}^{k\times3}$ and then either \textbf{(a)} stacks these transformed points in a single $\mathbb{R}^{3k}$ vector or \textbf{(b)} keeps the transformed points as $\mathbb{R}^{k\times3}$. 
In \textbf{(a)}, the original rigid transformation in $\mathbb{R}^{3}$ is not equivalent to a rigid body transformation in this $\mathbb{R}^{3k}$ space. Therefore, the 
embedding in the corresponding $\mathbb{ME}^{3k}\cong\mathbb{R}^{3k+2}$ conformal space will not be injective under 3D rigid body transformations --- the embedding will be \textit{invariant} to a \textit{too large} class of transformations. Whereas in \textbf{(b)}, one embeds these transformed $\mathbb{R}^{k\times3}$ points point-wise in $\mathbb{ME}^{3}\cong\mathbb{R}^5$, and the transformation maps one-to-one onto the conformal space (resulting in $\mathbb{R}^{k\times5}$).

We perform this point-wise embedding only in the first layer, but it can be done in the remaining layers as well.
In order to propagate the embedded input through the initial linear layer, we vectorize the $\mathbb{R}^{k\times5}$ array row-wise into $\textbf{\textit{X}}\in \mathbb{R}^{5k}$. For the proof of concept, we 
assume that the points are ordered. The case of point sets, i.e., including permutations, can be addressed by heuristics or max-pooling over permutations, but is not further considered here. 
Each intermediate layer output is a one-dimensional array, $\textbf{z} \in \mathbb{R}^{m}$, that we embed in $\mathbb{R}^{m+2}$, the same as in MLHP.
We illustrate the embeddings in Fig.~\ref{fig:model}.

We discover that this change of the embedding scheme affects the decision surfaces of the corresponding layer units.
Namely, for a given (embedded and vectorized) input point cloud $\textbf{\textit{X}}\in \mathbb{R}^{5k}$, a single unit in the first layer with weights $\widetilde{\textbf{\textit{S}}} \in \mathbb{R}^{5k}$  represents $k$ hyperspheres: one for each 3D point in the original $\mathbb{R}^{k\times3}$ input.
We call such units \textit{geometric neurons} (for a mathematical formulation, see Section~\ref{learned_param_interpret}) and our model the \textit{multilayer geometric perceptron} (MLGP).
Note that the proposed method works for any dimension other than three and, given a single 3D point as input, the first layer unit is identical to the hypersphere neuron \cite{banarer2003hypersphere}.

The embedding, which is non-linear and present at each layer, may eliminate the need for activation functions implied by MLPs. 
The choice of the final layer activation function depends solely on the application and is no different from the standard MLP case.

\subsection{Learned parameter interpretation}
\label{learned_param_interpret}
Since we regard the model parameters as independent during training, as proposed by \cite{banarer2003hypersphere}, our model learns \textit{non-normalized} hyperspheres (parameter vectors) of the form $\widetilde{\textbf{\textit{S}}} = ({s_1}, {s_2}, \dots, {s_{n+2}}) \in \mathbb{R}^{n+2}$.
We recall that due to the homogeneity of the hypersphere representation  (\ref{homogeneous_embedding}), both normalized and non-normalized hyperspheres represent the same decision surface. 

To analyze the learned decision surfaces in the respective Euclidean space, we need to obtain normalized vectors $\textbf{\textit{S}}$ as in (\ref{hypersphere_in_r}). To achieve this, we perform \emph{point normalization}, \ie, divide all elements in the learned parameter vector $\widetilde{\textbf{\textit{S}}}$ by the last one. We refer to this last element, ${s_{n+2}}$, as the \textit{scale factor}, $\gamma \in \mathbb{R}$. The scale factors can take arbitrary values. 

As a result, after training, we can alternatively decompose the geometric neuron output as a weighted sum of the scalar products of $k$ embedded input points and $k$ learned hyperspheres that the neuron represents and that are normalized. This decomposition allows interpreting the decision-making process in the Euclidean space. To fully appreciate this mental picture, the reader is encouraged to refer to Fig.~\ref{fig:model}.

Thus, we define the geometric neuron mathematically as
 \begin{equation}
	\label{geometric_unit_output}
	z = \sum_{i=1}^{k}\gamma_i \;  \textbf{\textit{X}}_i^\top \textbf{\textit{S}}_{i}\;,
\end{equation}
where $z \in \mathbb{R}$ is the output of the geometric neuron, $\text{\textbf{\textit{X}}}_i \in \mathbb{R}^{5}$ is the $i^\textup{th}$ row in the point-wise embedded input $\textbf{X} \in \mathbb{R}^{k\times5}$, and $\textbf{\textit{S}}_i = \widetilde{\textbf{\textit{S}}}_i / \gamma_i \in \mathbb{R}^{5}$ are the corresponding normalized learned parameters (hyperspheres). Each  $\text{\textbf{\textit{X}}}_i$ and $\text{\textbf{\textit{S}}}_i$ are of the form shown by (\ref{hypersphere_in_r}).

\begin{figure}	
	\includegraphics[width=1\linewidth]{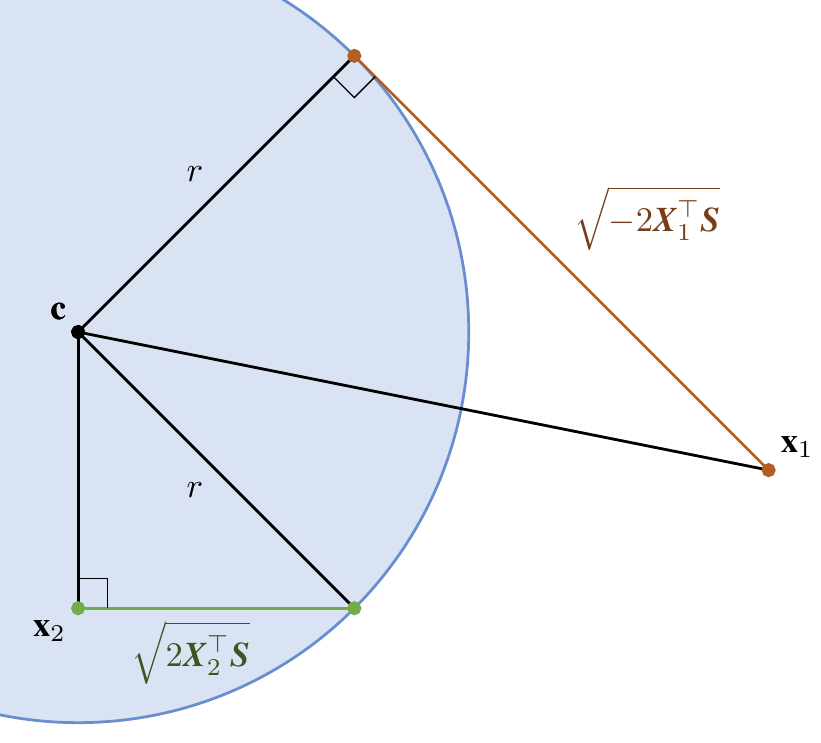}	
	\caption{The 2D interpretation of the scalar product of a point $X$ and a sphere $S$ in the conformal space (\ref{eq:scalar_product_in_conformal_space_euclidean_interpretation}). $\textbf{\textit{X}}_1$ and $\textbf{\textit{X}}_2$ are the Euclidean points $\textbf{x}_1$ and $\textbf{x}_2$ embedded in $\mathbb{R}^4\cong\mathbb{ME}^2$, respectively, and $\textbf{\textit{S}}$ is the conformal representation of the circle centered at $\textbf{c}$  and with radius $r$, described by (\ref{eq:sphere_in_conformal_space}). The scalar product $\textbf{\textit{X}}^\top \textbf{\textit{S}}$ determines the cathetus length.}
	\label{fig:scalar_product_intuition}
\end{figure}

We discuss the important effect of having negative scale factors in a later section. 
Also, radii of hyperspheres can be extracted from their normalized form (\ref{hypersphere_in_r}), namely from the second last element. 
However, as all other parameters, this element can be learned freely. It can even become negative, representing a hypersphere with an imaginary radius.  Although lacking geometric interpretation, this can be beneficial for the learning process \cite{banarer2003hypersphere}.

\subsection{Activation isometry in 3D}
\label{geometric_neuron_isometry}
An exciting property of the geometric neuron is that the rigid body transformations are isometries for its activations. There are two ways to show this.

First, consider transformations in the conformal space. We refer to the summary of motion operators and related conformal geometric algebra computations provided in \cite{hitzer2013geometric}.

Consider a motor $M$ (equation (35) in \cite{hitzer2013geometric}), \ie, rotation followed by translation in the conformal space,  operating on a general conformal object $O$, \eg, a point $X$ or a sphere $S$. To describe this transformation, we need to use the \textit{sandwich product}: 
\begin{equation}
	\label{eq:sandwich_product}
	O' = M^{-1} O M ~,
\end{equation}
where $M^{-1}$ is the inverse of the motor such that  $M^{-1}M = MM^{-1} = 1$, and $O'$ is the transformed object.

Suppose we now want to apply a motor $M$ to both a point $X$ and a sphere $S$ in the conformal $\mathbb{ME}^3$ space and compute their scalar product (\ref{eq:scalar_product_in_conformal_space_euclidean_interpretation}). By plugging \eqref{eq:sandwich_product} into \eqref{eq:inner},
we get
\begin{small}
\begin{equation}
	\label{eq:isometry_in_conformal_space}
	\begin{aligned}
		X' \cdot S' &= \frac{X'S' + S'X'}{2} \\
		&= \frac{M^{-1} X M \, M^{-1} S M + M^{-1} S M \, M^{-1} X M}{2} \\ 
		&= \frac{M^{-1} X S M + M^{-1} S X M}{2} \\ 
		&= M^{-1}\frac{X S + S X}{2} M = X \cdot S ~.
	\end{aligned}
\end{equation}
\end{small}

Given a rigid body transformation defined by the rotation $\textbf{R} \in \text{SO}(3)$ and translation $\textbf{t} \in \mathbb{R}^3$, the sandwich product (\ref{eq:sandwich_product}) of the corresponding motor $M$ and a sphere $S$ in the conformal $\mathbb{ME}^3$ space is isomorphic to the matrix-vector product with the matrix operator $\textbf{\textit{M}}_S$ on $\mathbb{R}^5$ defined as
\begin{small}
\begin{equation}
\label{eq:motor_isomorphism}
\textbf{\textit{M}}_S = 	
\begin{bmatrix}
\textbf{R} & ~\textbf{0} & ~\textbf{t}\\
\textbf{t}^\top\textbf{R} & 1 & \frac{1}{2}\textbf{t}^2\\
~\,\textbf{0}^\top & 0 & 1
\end{bmatrix} ~.
\end{equation}
\end{small}
The correctness of the isomorphism follows from computing 
\begin{small}
\begin{equation}
\textbf{\textit{M}}_S\textbf{\textit{S}}=\begin{bmatrix}
\textbf{R}\textbf{c}+\textbf{t} \\
\textbf{t}^\top\textbf{R}\textbf{c}+\frac{1}{2}(\textbf{c}^2-r^2)+ \frac{1}{2}\textbf{t}^2\\
1
\end{bmatrix}=\begin{bmatrix}
\textbf{R}\textbf{c}+\textbf{t} \\
\frac{1}{2}((\textbf{R}\textbf{c}+\textbf{t})^2-r^2)\\
1
\end{bmatrix} ~.
\end{equation}
\end{small}

To the best of our knowledge, this result has not been presented in any related work.

For the point $X$ embedded in the conformal $\mathbb{ME}^3$  space as shown in (\ref{hypersphere_in_r}), the corresponding operator $\textbf{\textit{M}}_X$ on $\mathbb{R}^5$ differs slightly, as elements 4 and 5 are swapped and negated
\begin{small}
\begin{equation}
\label{eq:motor_isomorphism2}
\textbf{\textit{M}}_X = 	
\begin{bmatrix}
\textbf{R} & -\textbf{t} & \textbf{0}\\
~ \,\textbf{0}^\top & 1 & 0\\
-\textbf{t}^\top\textbf{R} & \frac{1}{2}\textbf{t}^2 & 1
\end{bmatrix} ~,
\end{equation}
\end{small}
such that $\textbf{\textit{M}}_X$ is the adjoint of $\textbf{\textit{M}}_S$, \ie, $\textbf{\textit{M}}_X^\top \textbf{\textit{M}}_S=\textbf{\textit{I}}_5$.

Recalling the scalar product isomorphism between $\mathbb{ME}^n$ and   $\mathbb{R}^{n+2}$ (\ref{scalar_products_connection}) and using the result (\ref{eq:isometry_in_conformal_space}), we can see that applying the isomorphic operators $\textbf{\textit{M}}_X$ and $\textbf{\textit{M}}_S$ to both each point $\text{\textbf{\textit{X}}}_i$ and the respective sphere $\text{\textbf{\textit{S}}}_i$ in (\ref{geometric_unit_output}) does not change the output of the geometric neuron since
\begin{equation}
\label{eq:geometric_neuron_isometry}
(\textbf{\textit{M}}_X \textbf{\textit{X}})^\top (\textbf{\textit{M}}_S \textbf{\textit{S}}) = X' \cdot S' = X \cdot S = \textbf{\textit{X}}^{\top} \textbf{\textit{S}}=\textbf{\textit{X}}^\top\textbf{\textit{M}}_X^\top  \textbf{\textit{M}}_S \textbf{\textit{S}} ~. 
\end{equation}

Another way to show this is to recall the Euclidean space interpretation of the scalar product in the conformal space (\ref{eq:scalar_product_in_conformal_space_euclidean_interpretation}), which we illustrate by Fig.~\ref{fig:scalar_product_intuition} for the 2D geometry. Considering the cases when a point is either inside or outside the circle, we can construct the triangles accordingly. Applying rigid body transformations to both the circle and the point will preserve the triangles and hence the distance determined by the scalar product $\textbf{\textit{X}}^\top \textbf{\textit{S}}$ (i.e., the cathetus length). Thus, the geometric neuron decomposed as the sum of the scalar products (\ref{geometric_unit_output}) is indeed isometric in the corresponding Euclidean space.
We are not aware of any such graphical explanation previously presented in the literature.

In the following sections, we experimentally demonstrate this property and discuss its implications.

\section{Experiments}
\label{experiments}
To demonstrate the explainability of our method, we test it on a geometry classification problem. We also compare its performance with those of analogous baseline MLHP and vanilla MLP.
\subsection{3D shape classification data}
\label{shape_classification}
\begin{figure}
	\centering
	\includegraphics[width=0.9\linewidth]{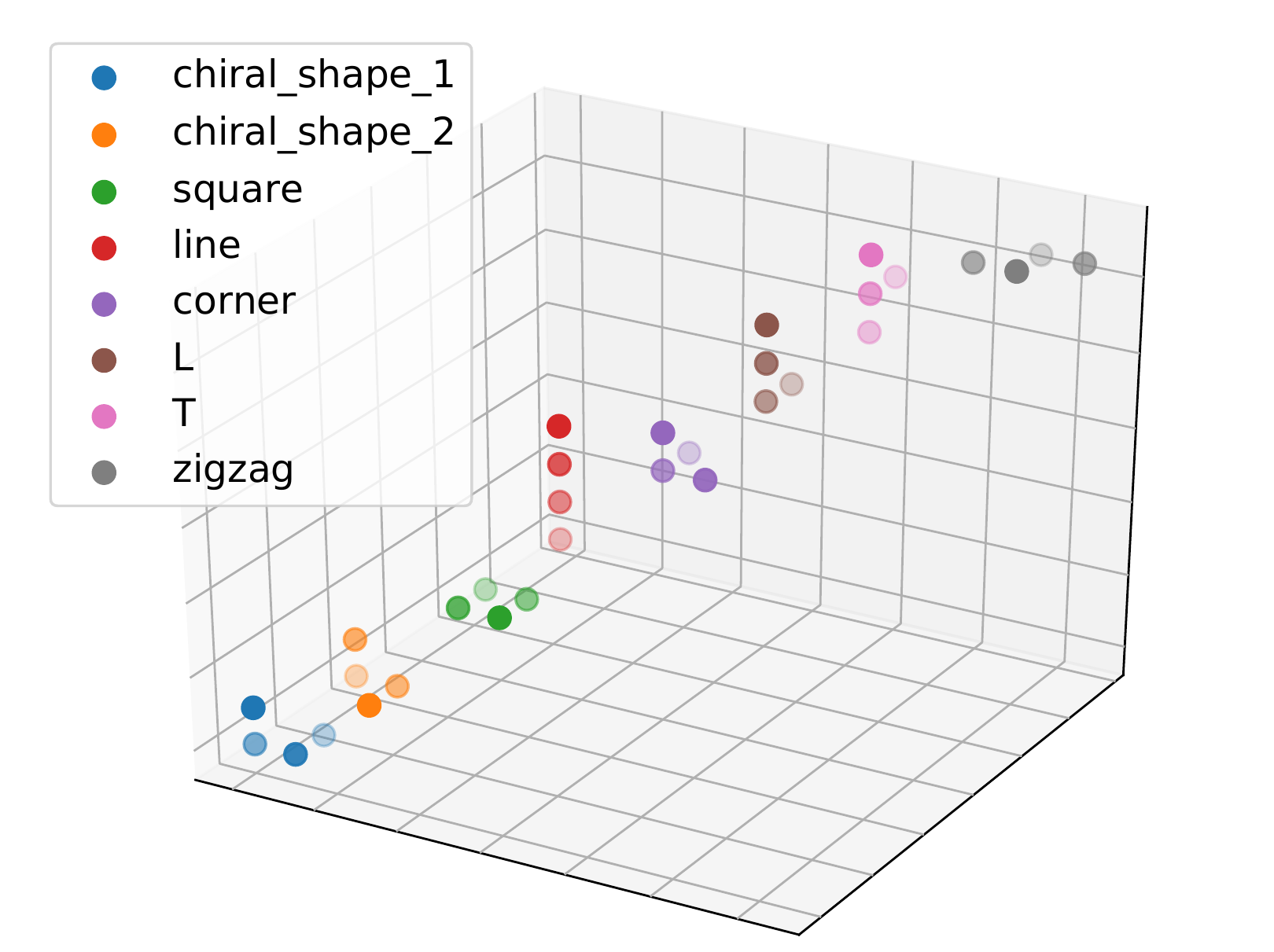}
	\caption{The 3D Tetris data.}
	\label{fig:tetris}
\end{figure}
We use the 3D Tetris dataset proposed in \cite{thomas2018tensor}. It consists of 8 shapes, displayed in Fig.~\ref{fig:tetris}. Each data sample is a $4 \times 3$ array, containing the 3D coordinates of four points in a certain order. We refer to these 3D coordinates as canonical. Note that the dataset includes two chiral shapes that are reflections of each other.

\textbf{Main dataset.\hspace{1em}}For the first experiment, we augment the Tetris data by performing uniform random rotation in $[0,2\pi)$ (about random axis) and translation in $(-3, 3)$, \ie, rigid body transformation of the canonical shapes. This way, we form a training set consisting of 1000 shapes, a validation set containing 9000 samples, and a test set of size 90000. 

\textbf{Theta-split.\hspace{1em}}Additionally, we create a \textit{theta-split} dataset of the same size to see if the models in our comparison can generalize over rigid transformations. The rotation angle, $\theta$, in the dataset construction differs for the training and validation/test sets: $\theta_\text{train}$ is drawn from the uniform distribution over the joint interval $\left[ 0, \frac{\pi}{4} \right) \cup \left[\pi, \frac{5\pi}{4} \right)$ and $\theta_\textup{val}$ and $\theta_\textup{test}$ from the antipodal interval. The translation vector is drawn as in the main dataset construction.

\textbf{Data with noise.\hspace{1em}}An important practical consideration for model comparison is that real-world data often contain a certain amount of noise. Therefore, we add  distortion, $\textbf{\textit{n}}$, of different levels to the shape coordinates in the main and theta-split datasets: $\textbf{\textit{n}} \sim U(-a, a)$ with $a \in \{0.1,\, 0.2\}$. We thus obtain four additional datasets. 

\begin{table*}
	\scriptsize
	\centering
	\caption{Model accuracies on the \textit{test} data (mean and std over 50 runs, \%); values in parentheses represent the accuracy of the 10 best models selected based on the \textit{validation} accuracy.}
	
	\label{tab:test_accs}
	\begin{tabular}{lccc@{}cccc}
		\toprule
		& \multicolumn{3}{c}{Main dataset} &&  \multicolumn{3}{c}{Theta-split}\\
		\cmidrule(l){2-4} \cmidrule(l){6-8}
		Noise & $a=0.0$ & $a=0.1$ & $a=0.2$ & & $a=0.0$ & $a=0.1$ & $a=0.2$ \\
		\midrule
		MLP   
		&$71.7\pm4.0$&$65.3\pm4.3$& $60.7\pm3.2$& &$55.3\pm5.8$&$51.8\pm3.4$&$48.4\pm3.6$\\
		
		&$(78.0\pm2.6)$&$(71.7\pm4.4)$& $(65.9\pm2.8)$& &$(63.8\pm3.1)$&$(56.5\pm2.2)$&$(53.3\pm1.9)$\\
		
		\addlinespace[0.5em]
		
		Baseline \cite{banarer2003design} 
		& $\textbf{92.5}\pm\textbf{0.4}$&$81.3\pm3.4$& $68.2\pm3.2$& &$87.6\pm0.6$&$78.4\pm3.8$& $65.7\pm3.7$\\
		
		& $(\textbf{92.8}\pm\textbf{0.2})$&$(86.9\pm0.4)$& $(72.2\pm0.5)$& &$(88.3\pm0.6)$&$(83.8\pm1.8)$& $(70.4\pm0.7)$\\
		
		\addlinespace[0.5em]

		Ours &$91.8\pm2.2$&$\textbf{81.8}\pm\textbf{5.9}$&$\textbf{69.7}\pm\textbf{8.3}$&&$\textbf{87.9}\pm\textbf{0.8}$&$\textbf{80.3}\pm\textbf{5.1}$&$\textbf{68.2}\pm\textbf{8.4}$\\
		
		&$(92.5\pm0.2)$&$(\textbf{89.3}\pm\textbf{0.4})$&$(\textbf{79.6}\pm\textbf{0.2})$&&$(\textbf{89.0}\pm\textbf{0.4})$&$(\textbf{87.5}\pm\textbf{0.5})$&$(\textbf{79.4}\pm\textbf{0.3})$\\
		
		\bottomrule
	\end{tabular}	
\end{table*}

\subsection{Setup}
When building models for the experiments, we want the total number of parameters to be comparable, even though a quantitative comparison of the methods is beyond the main focus of our work. The decision surfaces in our (MLGP) and the baseline (MLHP) models are of a higher order of complexity than the vanilla MLP case. As a consequence, they have a different number of hidden units. We select the vanilla model with 6 hidden units (134 parameters), the baseline MLHP model with 5 hidden units (126 parameters), and our MLGP with 4 hidden units (128 parameters). Note that the vanilla model includes bias parameters, whereas the other two do not.

We try different activation functions for all models: the sigmoid, hyperbolic tangent (tanh), ReLU, and identity, \ie, no activation function. In the case of MLGP and MLHP, identity means that the only source of non-linearity is the embedding.
The final layer of all models in our experiments is equipped with the softmax activation function.
We implement all models in PyTorch \cite{paszke2019pytorch} and use the default parameter initialization for linear layers. We train the models for 20000 epochs by minimizing the cross-entropy loss function with the Adam optimizer \cite{kingma2014adam} supplied with the default hyperparameters: the learning rate is set to $0.001$, $\beta_1 = 0.9$, and $\beta_2 = 0.999$.
We run each experiment 50 times. At each run, we generate the datasets (training and validation sets) described in Section~\ref{shape_classification}. The test data are generated once for each experiment.

We train and test the models on the main and theta-split datasets, both with different levels of noise. Since the three vanilla models with the identity, sigmoid, and tanh activation functions, respectively, are inferior to that with ReLU, we show only the latter case and proceed with ReLU as an activation function for the vanilla model. 
Our MLGP method and the baseline model perform much better without activation functions, which motivates us to use this configuration. 
The performances of the models on the test data and in all experiments are presented in Table~\ref{tab:test_accs}.

\subsection{Isometry test}
\label{isometry_test} 
To show that the rigid body transformations are isometries for the geometric neuron activations as claimed in Section~\ref{geometric_neuron_isometry}, we design the following experiment. Given an instance of our MLGP model pretrained on the main data (see Section~\ref{shape_classification}), we generate a rigid body transformation in $\mathbb{R}^3$ and apply it to the weights of the geometric neurons in the model by means of the isomorphism (\ref{eq:motor_isomorphism}). This results in a \textit{transformed} model. Subsequently, we compare the performance of the original model on the original test set and the performance of the transformed model on the test set modified by the generated transformation. Our hypothesis is that the two results will be indistinguishable up to a numerical precision if the geometric neuron activations are indeed isometric on $\mathbb{R}^3$. Table~\ref{tab:isometry_test} summarizes the experiment outcome. For the sake of completeness, we also evaluate the original model on the transformed test set and the transformed model on the original test set.

\begin{table}
    \small
	\centering
	\caption{Isometry test: the pretrained (original) and transformed MLGP accuracies on the original and transformed \textit{test} split from the main dataset (mean and std over 10000 runs, \%).}
	\label{tab:isometry_test}
	\begin{tabular}{lcc@{}c}
			\toprule
			& MLGP && Transformed MLGP\\			
			\midrule 			
			
			Original data  
			&$\textbf{91.98}\pm\textbf{0.00}$ && $86.11\pm3.46$\\
		
			Transformed data
			&$86.09 \pm 3.45$ && $\textbf{91.98}\pm\textbf{0.00}$\\
		
		\bottomrule
	\end{tabular}	
\end{table}

\section{Discussion and conclusion}
\label{discussion}
In this section, we discuss the major benefits of the proposed MLGP model: the exlplainability of coefficients and the improved quantitative results, in particular when extrapolating transformation parameters.

\subsection{Model explainability}
\label{conformal_model_interior}
One of the main advantages of the embedding scheme utilized in our  model is that it provides an intuitive geometrical interpretation of the learned coefficients.
To visualize the learned decision surfaces in the Euclidean $\mathbb{R}^3$ space, we need to point-normalize them according to (\ref{hypersphere_in_r}).

\begin{figure}
	\centering
	\includegraphics[width=1\linewidth]{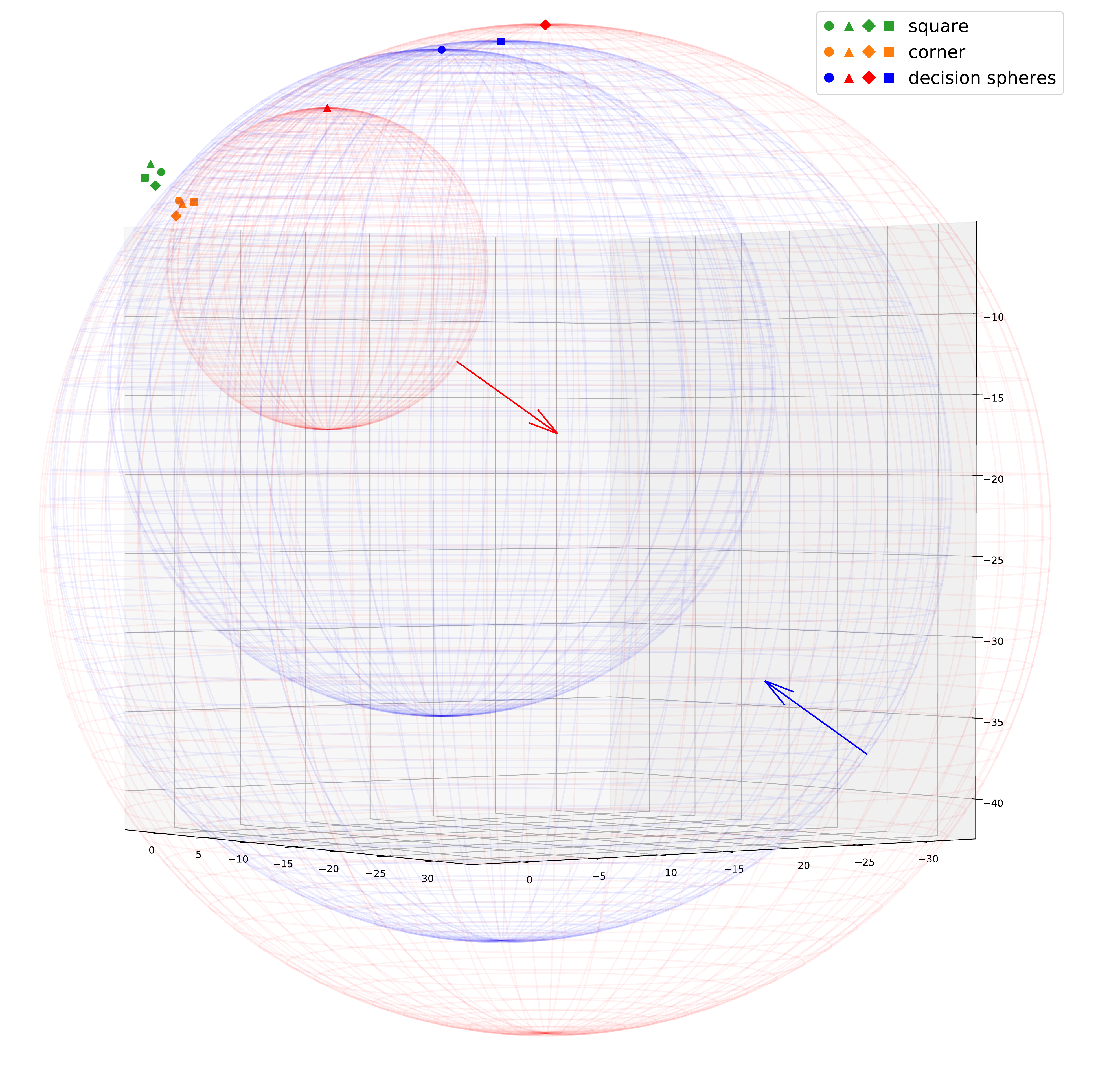}\\[-2ex]
	\includegraphics[width=1\linewidth,trim={0 2cm 0 0cm},clip]{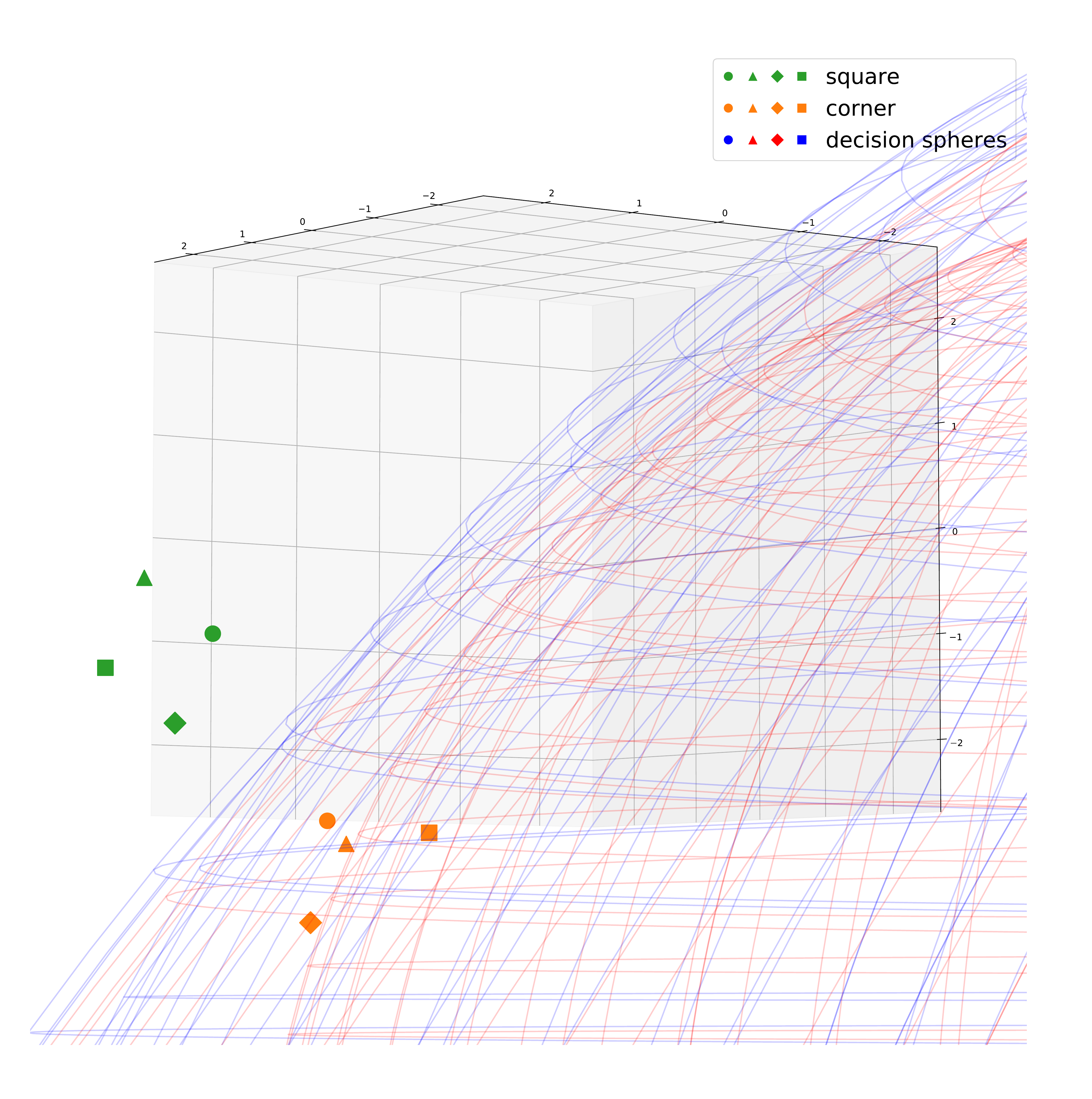}
	\caption{A single unit in the first layer (i.e., geometric neuron) in our MLGP model classifying the Tetris shapes: each spherical decision surface classifies the corresponding point of the input shape (specified by the markers); the unit output is then a linear combination of the scalar products (\ref{geometric_unit_output}). Top: the arrows specify the positive direction of the scalar product, \ie, inside or outside the sphere. Bottom: a zoomed-in view.}
	\label{fig:norm_hyps}
\end{figure}

\begin{figure}
	\includegraphics[width=0.9\linewidth,trim={2cm 5cm 0 0},clip]{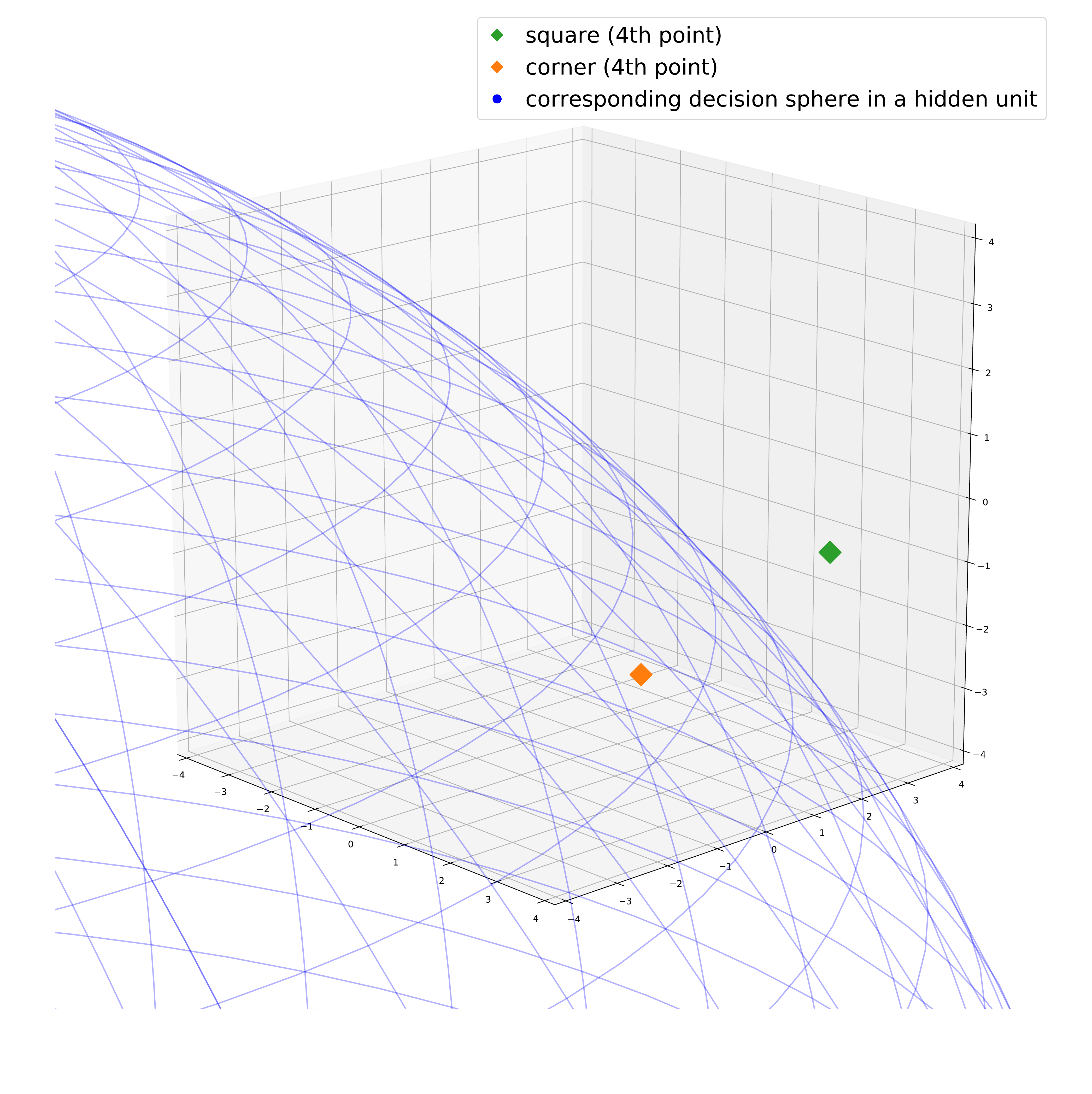}	
	\caption{One of the spheres in a single geometric neuron; the sphere classifies the corresponding point in the input shapes: the scalar product of one point of the \textit{corner} and the displayed normalized sphere $\approx 39.15$ (positive), whereas that of the point of the \textit{square} is  $\approx -50.75$ (negative).}
	\label{fig:one_norm_hyp}
\end{figure}

We use two shapes from the main dataset described in Section~\ref{shape_classification} as input to the trained MLGP model. We demonstrate the input shapes and the four spherical decision surfaces represented by the (fourth) hidden unit in Fig.~\ref{fig:norm_hyps}, wherein each spherical decision surface classifies the corresponding point of the input shape. For clarity, we show how the fourth decomposed sphere of a single hidden unit in the same model distinguishes the respective points of the two input shapes in Fig.~\ref{fig:one_norm_hyp}.

Note that each sphere may have a different scale factor. It can be negative and, thanks to the normalization step, turn  the decision surface inside out for a given input. This phenomenon is discussed but not visualized in the prior work. Importantly, this swap is itself a conformal transformation. 
Suppose the sign of the scalar product (\ref{scalar_products_connection}) of an embedded input point \textbf{\textit{X}} and a sphere \textbf{\textit{S}} is the same regardless of the normalization of the latter. In that case, the input is categorized as class $\mathcal{I}$ if it is inside or on the surface of the sphere, and to class $\mathcal{O}$ if it is outside. We refer to such hyperspherical classifiers as $\mathcal{I}$\textit{-hyperspheres} and $\mathcal{O}$\textit{-hyperspheres}, respectively.
We illustrate the idea of the inverted decision surfaces by drawing $\mathcal{O}$-spheres in red, whereas $\mathcal{I}$-spheres are shown in blue (see Fig.~\ref{fig:norm_hyps} and Fig.~\ref{fig:one_norm_hyp}).

\subsection{Quantitative results}
The superiority of our MLGP model with no activation function other than the embedding and even fewer number of parameters (128 vs.$\,$134) to the plain MLP is evident from Table~\ref{tab:test_accs}.  
This result confirms the intuition from Section~\ref{method}, given that our model units have higher-order decision surfaces. However, it comes at the cost of increased computational complexity. Considering that we have to evaluate the magnitude of $m$ vectors at each layer in the embedding step, we can roughly compare it to adding $m$ extra neurons to the respective layer in an analogous vanilla MLP, in accordance with the complexity analysis given in \cite{banarer2003hypersphere}. 
Compared to the MLP, our model has an increased computational complexity with a factor of typically between 1.6 and 2.

What is remarkable is the embedding being non-linear and present at each layer allows for successful learning without any activation function. To the best of our knowledge, this has not been observed in prior work.

In all noisy data experiments, our MLGP demonstrates, on average, better generalization than the baseline and vanilla models (see Table~\ref{tab:test_accs}). 
We noted variations of the validation accuracy,  presumably due to confusing two or three classes, and selected the ten models of each type with best validation accuracy. Our method has the best correlation of validation accuracy and test accuracy, which further increases its advantage and reduces its variance.

We notice a major drop in the generalization performance of the vanilla MLP in the case of the theta-split data
and a smaller, yet significant, drop for the baseline, whereas the generalization accuracy of our method decreases insignificantly, as indicated by Table~\ref{tab:test_accs}. 
To a certain extent, this suggests that hyperspherical decision surfaces are better suited for such geometry tasks than standard hyperplanes. This has not been discussed in prior work.

Overall, the experiments show that in addition to being more geometrically motivated and explainable, our modification to the baseline MLHP, the MLGP, produces favorable quantitative results classifying the 3D Tetris shapes and is even superior when they are perturbed.

\subsection{Isometry implications}
The result shown in Section~\ref{geometric_neuron_isometry} is verified in Section~\ref{isometry_test}: when the geometric neuron parameters are transformed the same way as the input, the neuron activations remain unchanged, hence the model output and the performance of the model, as presented in Table~\ref{tab:isometry_test}. Although this result may appear mathematically trivial, its beauty is that we can apply it to a very complex set of model parameters since we know they follow the geometrically correct algebraic structure (\ref{geometric_unit_output}) by construction with the proposed embedding scheme in our MLGP method. Needless to say, the baseline MLHP input embedding would not allow for such manipulations simply because the $\mathbb{R}^3$ transformations of the input shapes would not be applicable to the $\mathbb{R}^{12}$ hyperspheres. The same argument applies to the vanilla MLP, since a hyperplane can be seen as a special case of a hypersphere (\ie, with infinite radius) \cite{perwass2003spherical}, and therefore, the MLHP is a generalization of the standard perceptron model.

The isometry property of the geometric neuron is a necessary condition to consider rotation and translation equivariance (discussed in, \eg, \cite{thomas2018tensor}) capabilities of our model. This is, however, outside the scope of this paper.

{
\section*{Acknowledgments}
	This work was supported by the Wallenberg AI, Autonomous Systems and Software Program (WASP), by the Swedish Research Council through a grant for the project Algebraically Constrained Convolutional Networks for Sparse Image Data (2018-04673), and the strategic research environment ELLIIT.
}

{\small
\bibliographystyle{ieee_fullname}
\bibliography{egbib}
}

\end{document}